\documentclass[10pt,twocolumn,letterpaper]{article}

\usepackage{iccv}
\usepackage{times}
\usepackage{epsfig}
\usepackage{graphicx}
\usepackage{amsmath}
\usepackage{amssymb}
\usepackage{multirow}
\usepackage{caption}
\usepackage[inkscapelatex=false]{svg}
\usepackage{paralist}
\usepackage{threeparttable}
\usepackage{url}
\usepackage[accsupp]{axessibility}

\usepackage{booktabs}
\usepackage{multicol}
\usepackage{multirow}
\usepackage{bm}
\usepackage{diagbox}
\usepackage{makecell}

\usepackage[pagebackref=true,breaklinks=true,letterpaper=true,colorlinks,bookmarks=false]{hyperref}
\iccvfinalcopy 

\newcommand{\figref}[1]{Figure~\ref{#1}}%
\newcommand{\tabref}[1]{Table~\ref{#1}}%

\renewcommand{\eqref}[1]{Eq.~(\ref{#1})}

\def\iccvPaperID{10310} 

\ificcvfinal\fi

\newif\ifdrafting
\draftingtrue 
\ifdrafting
    \newcommand{\todo}[1]{{\leavevmode\color[rgb]{0,0,1}[TODO: #1]}}
    \newcommand{\ds}[1]{{\leavevmode\color[rgb]{1,0,0}[Deqing: #1]}}
\else
	\newcommand{\todo}[1]{}
    \newcommand{\ds}[1]{}
\fi

\begin{document}

\title{SAMPLING: Scene-adaptive Hierarchical Multiplane Images Representation \\ for Novel View Synthesis from a Single Image}

\author{
Xiaoyu Zhou\textsuperscript{1}
~ Zhiwei Lin\textsuperscript{1}
~ Xiaojun Shan\textsuperscript{1}
\thanks{Work is done during the internship at Peking University.}
~ Yongtao Wang\textsuperscript{1}
\thanks{Corresponding author.}
~ Deqing Sun\textsuperscript{2}
~ Ming-Hsuan Yang\textsuperscript{2,3} \\
{\textsuperscript{1}Wangxuan Institute of Computer Technology, Peking University}\\ 
{\textsuperscript{2}Google Research}
~ {\textsuperscript{3}University of California, Merced} \\
}


\maketitle
\ificcvfinal\thispagestyle{empty}\fi

\begin{abstract}
%
Recent novel view synthesis methods obtain promising results for relatively small scenes, \eg, indoor environments and scenes with a few objects,  
but tend to fail for unbounded outdoor scenes with a single image as input. 
In this paper, we introduce SAMPLING,  a \textbf{S}cene-\textbf{a}daptive Hierarchical \textbf{M}ulti\textbf{pl}ane \textbf{I}mages Representation for \textbf{N}ovel View Synthesis from a Sin\textbf{g}le Image based on improved multiplane images (MPI).
%
Observing that depth distribution varies significantly for unbounded outdoor scenes, we employ an adaptive-bins strategy for MPI to arrange planes in accordance with each scene image.
%
To represent intricate geometry and multi-scale details, we further introduce a hierarchical refinement branch, which results in high-quality synthesized novel views.
Our method demonstrates considerable performance gains in synthesizing large-scale unbounded outdoor scenes using a single image on the KITTI dataset and generalizes well to the unseen Tanks and Temples dataset.
%
%
The code and models will be made available at \url{https://pkuvdig.github.io/SAMPLING/}.
\end{abstract}


\section{Introduction}
Taking a photo and using it to synthesize photo-realistic images at novel views is an important task with a wide range of applications, such as generating realistic data for training AI models (\eg, autonomous driving perception and robot simulation). 
This task is challenging as it requires a precise understanding of 3D geometry, reasoning about occlusions, and rendering high-quality, spatially consistent novel views from a single image.
It becomes even more difficult for large-scale unbounded outdoor scenes, which contain complex geometric conditions, various objects, and diverse depth distributions corresponding to different scenes.

%
Recently, Neural Radiance Field (NeRF)~\cite{mildenhall2021nerf} based methods have gained much attention by synthesizing photo-realistic images with dense multi-view inputs. 
By leveraging Multi-layer Perceptron (MLP) layers, NeRF implicitly models a specific scene via RGB values and volume occupancy density. 
However, NeRF-based methods are primarily applicable for rendering bounded objects or interiors, which are impeded by the stringent requirement for the dense views captured from different angles, precise corresponding camera poses, and unobstructed conditions~\cite{barron2021mip, deng2022depth, martin2021nerf}.
Furthermore, these methods rely on per-scene fitting and cannot easily generalize to unseen scenes. 
Several methods~\cite{li2022read, rakhimov2022npbg++, carlson2022cloner, xie2023s} try to utilize multi-modal data, \textit{e.g.}, LiDAR scans and point clouds, to complicate the synthesis of novel views in large scenes.
However, additional modalities are difficult to obtain and have greater memory consumption and computational costs. 
%
%
Besides, similar to NeRF, these multi-modal methods require multiple input views with large overlaps and need to be trained per scene. 

In contrast, the Multiplane Images (MPI) representation~\cite{tucker2020single} has shown promising results in synthesizing scenes from sparse views, using a set of parallel semi-transparent planes to approximate the light field.
The MPI representation is particularly effective at understanding complex scenes with challenging occlusions~\cite{wizadwongsa2021nex}.
%
However, prior MPI-based approaches place planes at fixed depths with equal intervals, have limitations in modeling irregular geometry, such as texture details, and do not perform well in unbounded outdoor scenes, as shown in \figref{fig:teaser}. 
For complicated geographic features and differentiated depth ranges, the uniform static MPIs~\cite{zhou2018stereo, tucker2020single, li2021mine} are often over-parameterized for large areas of space, yet under-parameterized for the occupied scenes.
%
%
In addition, using single-scale scene representation in MPI also limits the quality of the synthesized images in large-scale scenes, leading to apparent artifacts and blurs.

In this paper, we introduce SAMPLING, a scene-adaptive hierarchical representation for novel view synthesis from a single image based on improved MPI. 
%
Instead of generating multiplanes with a static uniform strategy, we design the Adaptive-bins MPI Generation strategy to adaptively distribute the planes according to each input image.
This strategy enables a more efficient representation to better fit various unbounded outdoor scenes.
%
Additionally, we propose a Hierarchical Refinement Branch that utilizes multi-scale information from large scenes, incorporating both global geometries and high-frequency details into the MPI representation.
This branch enhances the quality of intermediate scene representations, resulting in more complete and high-quality synthesized images.
Our method achieves high-quality view synthesis results on challenging outdoor scenes, such as urban scenes, and shows a well cross-scene generalization, enabling a more versatile scene representation.
%
Our main contributions are:
\begin{compactitem}
    \item We present a novel scene-adaptive representation for synthesizing new views from a single image. Our approach is based on learnable adaptive-bins for MPI, enabling the learning of a more efficient and effective unbounded scene representation from a single view.
    \item We develop a hierarchical refinement method for 3D representation of outdoor scenes. We show that representing scenes with hierarchical information can synthesize new images with favorable details.
    \item Our method achieves new state-of-the-art performance in outdoor view synthesis from a single image.
    Experimental results also show our method generalizes well for both outdoor and indoor scenes.
\end{compactitem}

\begin{figure*}[htbp]
  \centering
  \includegraphics[width=\linewidth]{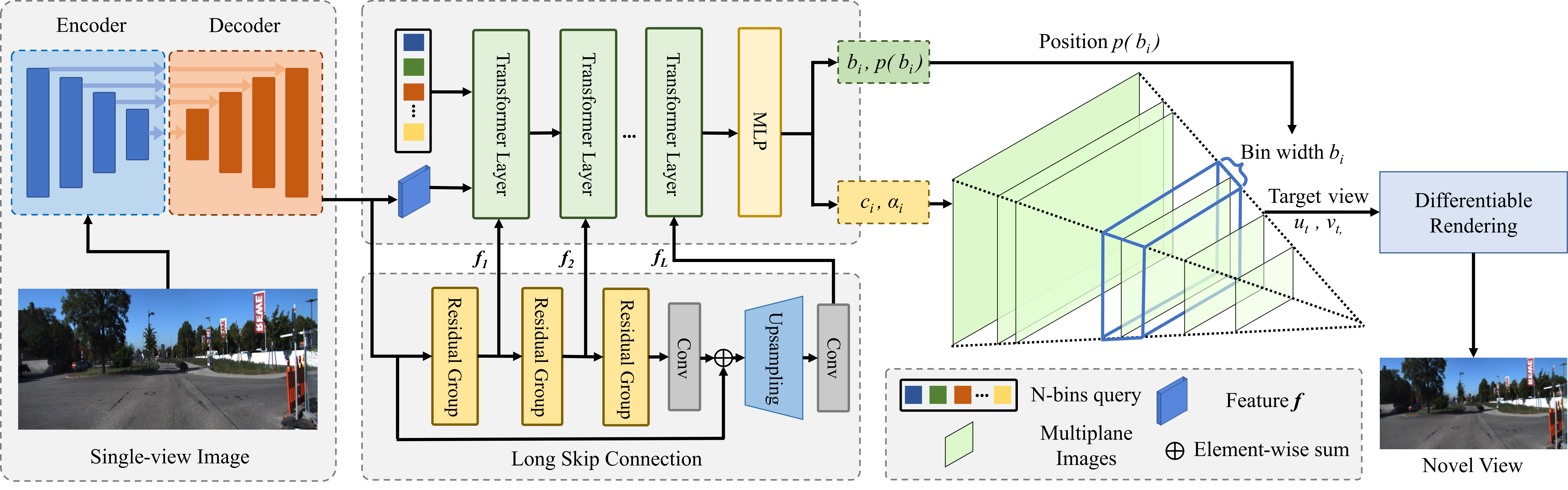}
  \caption{\textbf{An overall pipeline of our proposed method for novel view synthesis from a single image.} Given a single-view image as input, we first employ an encoder-decoder network combined with skip connections to extract features. The features are then fed into the Adaptive-bins MPI generation module along with an N-bins query, which calculates the adaptive positions of the MPI. Simultaneously, the Hierarchical Refinement Branch extracts hierarchical residual features with a set of Residual Groups and passes them to Transformer Layers. The MPI position $p(b_i)$ and representation $(c_i, \alpha_i)$ are then predicted by an MLP head to synthesize the novel views through the differentiable rendering.}\label{fig:overview}
  \vspace{-5pt}
\end{figure*}

\section{Related Work}
\subsection{Novel View Synthesis}
Novel view synthesis (NVS) aims to render unseen viewpoints of the scene from the observed images.
Recently, numerous deep models have been introduced to represent 3D objects or scenes and synthesize images in novel views.
%
%
Some methods exploit generative models for image generation and completion~\cite{chen2020generative, wiles2020synsin, tewari2022disentangled3d, gao2022get3d}, while others exploit explicit or implicit 3D scene representations~\cite{mildenhall2021nerf, wang2022r2l, tulsiani2018layer, zhou2018stereo} derived from input images and synthesize new viewpoints through differentiable rendering.
%


The recent methods based on the neural radiance field (NeRF)~\cite{pumarola2021d, barron2022mip, turki2022mega} have achieved state-of-the-art results for implicit neural 3D scene representation.
Given a set of posed images, NeRF methods map the 3D position and direction to a density and radiance by the multilayer perceptron (MLP), followed by differentiable volume rendering to synthesize the images. 
Typically, the original NeRF~\cite{mildenhall2021nerf} model is trained per scene and requires dense inputs with accurate camera poses. 
To make NeRF more practical, the NeRF in the wild method~\cite{martin2021nerf} requires only unstructured collections of in-the-wild photographs. 
In~\cite{lin2021barf, wang2021nerf, meng2021gnerf}, 
Lin \etal~train NeRF from imperfect (or even unknown) camera poses.
%
%
Other approaches explore the possibilities of NeRF in more application scenarios, such as dynamic scenes~\cite{pumarola2021d, wu2022d, xu2021h}, controlled editing~\cite{yuan2022nerf, wang2022clip}, and interior scenes~\cite{wei2021nerfingmvs, chen2022structnerf}. 
However, if the inputs are sparse (or even a single image), or the scene is large and complicated (\textit{e.g.}, urban street view), the novel views synthesized by NeRF-based methods will be of low quality and contain artifacts~\cite{yu2021pixelnerf}.
%
Furthermore, existing works for novel view synthesis need to be trained per scene, lacking general representation for scene understanding.

In contrast to NeRF, Multiplane Images (MPI) methods can synthesize novel views from fewer images, due to the properties of explicitly modeling scenes with sparse inputs. 
Using a stack of RGB-$\alpha$ layers at various depths, the MPI representation mimics the light field in a differentiable manner. 
In recent years, significant advances have been made in MPI for novel view synthesis. 
For instance, Zhou \etal~\cite{zhou2018stereo} use MPI for realistic rendering of novel views with a stereo image pair.
%
%
In~\cite{tucker2020single}, an MPI-based method is developed to synthesize views directly from a single image input, leading to higher-quality results compared to traditional light fields. 
DeepView~\cite{flynn2019deepview} further applies learned gradient descent to estimate multiplane images from sparse views, replacing the simple gradient descent update rule with a deep network. 
To improve the real-time performance, NeX~\cite{wizadwongsa2021nex} models view-dependent effects by performing basis expansion on the pixel representation.
MINE~\cite{li2021mine} takes advantage of the MPI and NeRF, proposing a continuous depth MPI method for NVS and depth estimation.
AdaMPI~\cite{han2022single} improves MPI by adjusting plane depth and predicting depth-aware color with the help of depth maps estimated by off-the-shelf monocular depth estimators.
%
%
However, these approaches have limitations in modeling unbounded outdoor scenes with multi-scale information and complex geometry. They also fail to obtain detailed high-frequency information, leading to apparent artifacts, blurs, and defects when synthesizing images in large-scale scenes.

\subsection{Large-Scale Neural Scene Rendering}
Recent advances in neural rendering have exhibited considerable success in 3D object modeling and interior scene reconstruction. 
Nevertheless, current methods demonstrate suboptimal performance when applied to unbounded outdoor scenes. 
Numerous methods have been developed to address this issue. 
Block-NeRF~\cite{tancik2022block} enables large-scale scene reconstruction by dividing large environments into multiple blocks and representing each block with an individual NeRF network.
BungeeNeRF~\cite{xiangli2022bungeenerf} introduces a progressive neural radiance field, which models diverse multi-scale scenes with varying views on multiple data sources. 
However, these methods can only model large outdoor driving scenes that are observed from dense input sensor views and precise camera poses. 
With high-speed shots, the outdoor driving scenes typically have very sparse viewpoints and limited view diversity. 
To tackle the above challenges, recent methods have explored multi-modal fusion methods for neural rendering. 
Rematas \etal~\cite{rematas2022urban} extend NeRF to leverage asynchronously captured LiDAR data and to supervise the density of rays.
Similarly, CLONeR~\cite{carlson2022cloner} introduces the camera-LiDAR fusion to the outdoor driving scene, where LiDAR is used to supervise geometry learning. 
Li \etal~\cite{li2022read, ruckert2022adop, aliev2020neural} propose to synthesize photo-realistic scenes with the help of large-scale point clouds, using neural point-based rendering.
%
However, current multi-modal approaches take a two-stage synthesis strategy, that is, first pre-processing all multi-modal data to reconstruct a rough 3D scene and then rendering a novel view image from the reconstructed 3D scene.
%
Costly multi-modal data collection, complex pre-processing, and per-scene training limit the efficiency and application of these methods.
In contrast, we introduce a high-efficient representation for novel view synthesis called SAMPLING.
With only a single image as the input, our method can generate novel view images from end to end and produce more realistic results with fewer artifacts and deformities for a wide range of real-world scenes.
Besides, our method does not necessitate per-scene optimization and thus reduces training costs.
%

\section{Method}
The overview architecture of SAMPLING is shown in \figref{fig:overview}.
%
Given a single image $I$, SAMPLING learns to generate the multiplane images (MPI) representation with discretized adaptive-bins and hierarchical feature refinement module. 
Synthesis image $\hat{I_t}$ can then be rendered at various novel viewing angles from the generated MPI.

\subsection{Adaptive-bins MPI Generation}

We utilize MPI to explicitly represent the 3D geometry of the source view.
MPI consists of $N$ front-parallel RGB-$\alpha$ planes arranged at depths $d_0, \ldots d_{N+1}$.
Each plane $i$ encodes an RGB color image $c_{i}$ and an alpha map $\alpha_{i}$.  

Most existing works employ a uniform-fix MPI distribution strategy (\textit{e.g.}, MPI~\cite{tucker2020single} and MINE~\cite{li2021mine}), where planes are placed at fixed depths with equal intervals. 
%
However, depth distribution corresponding to different RGB inputs can vary dramatically, especially for outdoor scenes. 
%
Thus, we introduce an adaptive binning strategy for MPI generation.
We discretize the depth interval into $N$ bins, where the bin widths are adaptively obtained for each image, and distribute each plane of MPI according to the adaptive bins.
%

\begin{figure}[h]
  \centering
  \includegraphics[width=8.3cm]{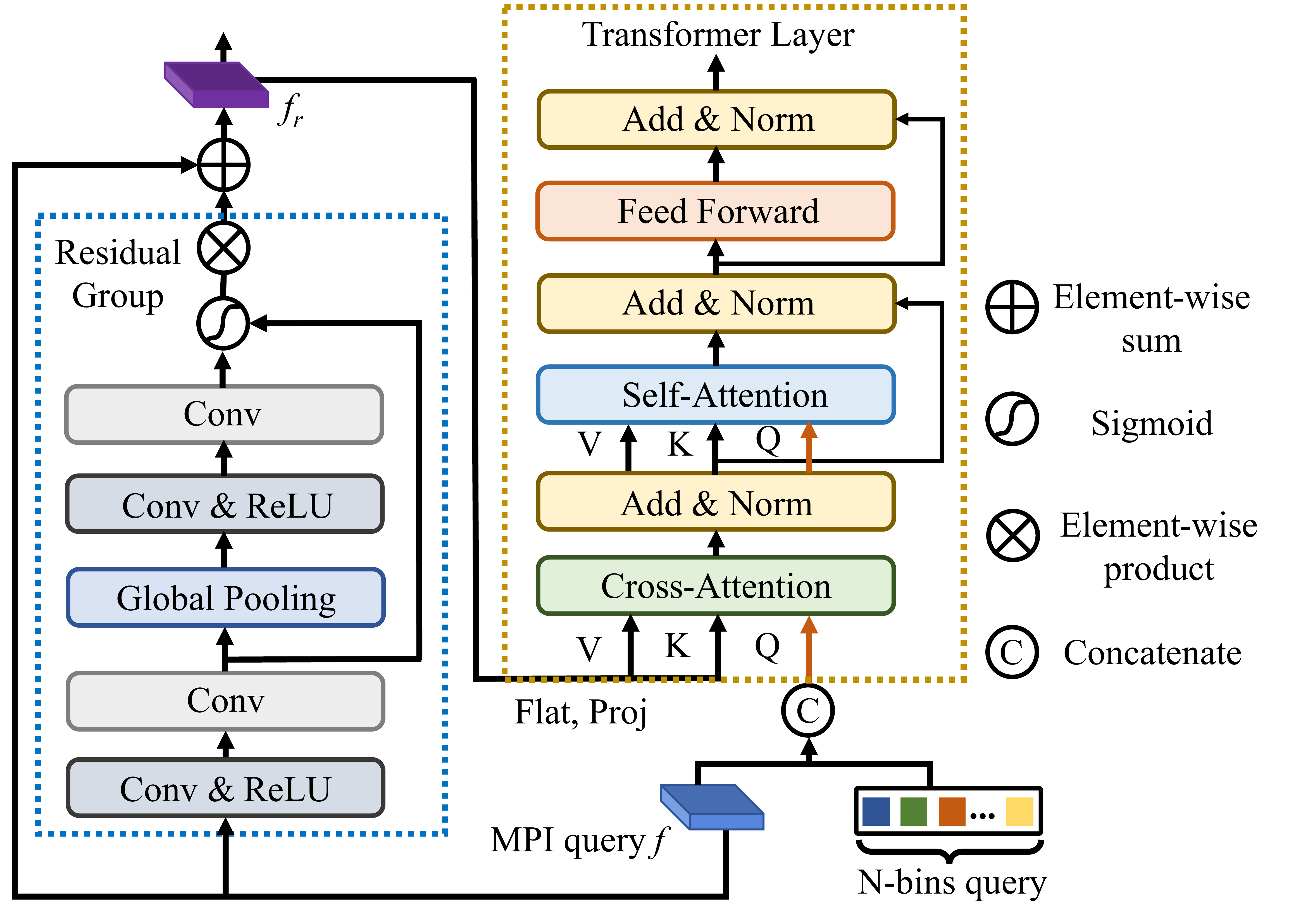}
  \caption{\textbf{Network details of the connection of Transformer Layer and Residual Group.} The combination of the two modules enables MPI representation to obtain both precise distribution and multi-scale detailed information.}\label{fig:transformer-layer}
  \vspace{-5pt}
\end{figure}

%
%
Specifically, we first extract the image feature $f$ by sending a single-view image into an encoder-decoder network. The encoder-decoder network utilizes skip connections to produce the high-resolution image feature in a coarse-to-fine style.
%
Then, we employ a transformer module to calculate the distribution of adaptive-bins MPI. 
The transformer module consists of several transformer layers, as shown in \figref{fig:overview}. 
%
Similar to Adabins~\cite{bhat2021adabins} and Binsformer~\cite{li2022binsformer}, we randomly initialize $N$ learnable bin queries $f_b$ for depth prediction. 
%
Meanwhile, the feature $f$ is viewed as the MPI query for RGB-$\alpha$ plane predictions in MPI.
In each transformer layer, the MPI query $f$ is sent to a Hierarchical Refinement Branch to produce the residual feature $f_r$. 
The residual feature $f_r$ is viewed as values and keys to calculate the cross-attention with the concatenated queries $f_b$ and $f$.
%
Then, the updated concatenated queries $f_b$ and $f$ are subsequently sent to a self-attention layer and a feed forward layer, as shown in \figref{fig:transformer-layer}.
After that, a shared multi-layer perception head is performed over the N-bins query $f_b$ and feature $f$ to predict bin width $\widetilde{b}$ and generate $(c_{i}, \alpha_{i})$ for each plane of MPI.
We apply the softmax function to normalize the sum of widths $\widetilde{b}_i$ to 1 as follows:
\begin{equation}
\label{bin-width}
\{b_{i}\}_{i=1}^{N} = Softmax(\{\widetilde{b}_{i}\}_{i=1}^{N}) \,,
\end{equation}
where $b_i$ is the $i^{th}$ normalized bin width.
%
Finally, we calculate the adaptive depth location of each plane in MPI by:
\begin{equation}
\label{MPI-position}
p(b_{i}) = d_{near} + (d_{far}-d_{near})(\frac{b_{i}}{2}+\sum_{j=1}^{i-1}b_{j}) \,,
\end{equation}
where $p(b_{i})$ is the position assigned to the $i^{th}$ adaptive-bins MPI. $d_{near}$ and $d_{far}$ are the nearest and farthest distances of the planes in the frustum of the camera, respectively.

\subsection{Hierarchical Refinement Branch}

%
Synthesizing novel views from a single image often faces difficulty in capturing multi-scale scene features, resulting in visually obvious holes and blurs. 
To address this issue, we propose a Hierarchical Refinement Branch to improve the feature with multi-scale information, which has proven effective in both 3D scene representation~\cite{rosinol2022nerf} and single image super-resolution~\cite{li2023hst, cai2022hipa} tasks. 
%

Specifically, we employ a coarse-to-fine architecture, where the low-resolution planes enforce the smoothness in scenes, and high-resolution planes refine the geometry details.
%
%
Given the shallow feature $f$ from the encoder-decoder network, we employ a set of residual groups (RG) \cite{zhang2018image, niu2020single} with upscale modules to extract hierarchical residual features $\{f_{r} \vert r=1,2,...,L\}$, which can be formulated as:
%
\begin{equation}
\label{intermediate}
f_r = H_{RG_{r}}(f_{r-1}),
\end{equation}
where $H_{RG_{r}}$ represents the $r^{th}$ residual group, $L$ is the number of the residual groups. 
RG aims to restore the high-frequency information and to extract rich edge and texture information of the outdoor scenes. The structural details of the RG are shown in \figref{fig:transformer-layer}.
%

Besides, to stabilize the training process, we introduce a long skip connection, an additional upsampling block, and two convolution layers when calculating the last high-resolution residual feature $f_L$. 
Subsequently, we feed the output $f_L$ into the last transformer layer, encouraging the generation of MPI to pay more attention to the informative details of scenes.
%

%

\subsection{Differentiable Rendering in MPI}
The synthesized MPI can be rendered in the target view by first warping each plane from the source viewpoint and then applying the composite operator to aggregate the warping results of each plane. The overall MPI rendering can be formulated as follows:
\begin{equation}
\label{mpi_render}
\hat{I_{t}} = O(W(C), W(A)) \,,
\end{equation}
where $\hat{I_t}$ denotes the synthesized image, $W$ is the homography warping function, and $O$ is the composite operator. $C = \{c_{1}, \ldots, c_{N}\}$ denotes the set of RGB channels and $A = \{\alpha_1, \ldots, \alpha_N\}$ is the corresponding alpha channel. 

We first employ the homography warping operation for the $i^{th}$ plane from the target to source view depending on the position $p(b_i)$ of each plane. Given the rotation matrix $R$, the translation matrix $t$ from the target to source view, and the intrinsic matrix $K_s$ and $K_t$ for source and target views, we can generate the synthesized target-view image through $W$ as follows:
\begin{equation}
\label{warping}
\left[ u_s, v_s, 1 \right]^{\top} \sim K_s(R - \frac{tn^{\top}}{p(b_i)})(K_t)^{-1} \left[ u_t, v_t, 1 \right]^{\top} \,,
\end{equation} 
where $\left[ u_s, v_s\right]$ and $\left[ u_t, v_t\right]$ are coordinates in the source and target views, respectively. $n$ is the norm vector of the $i^{th}$ plane at the position $p(b_i)$. 
The MPI representation of the target view can be obtained by warping each layer from the source viewpoint to the desired target viewpoint using ~\eqref{warping}, finding the corresponding pixel for each pixel in the target frame. The MPI representation under the target view $(c_{i}^{\prime}, \alpha_{i}^{\prime})$ can be defined as:
\begin{eqnarray}
    f(x)=
	\begin{cases}
	c_{i}^{\prime}(u_t,v_t) = c_{i}(u_s,v_s), \\
	\alpha_{i}^{\prime}(u_t,v_t) = \alpha_{i}(u_s,v_s).
	\end{cases}
\end{eqnarray} 
Finally, the synthesized target-view image can be then rendered via the compositing procedure~\cite{porter1984compositing} as follow:
\begin{equation}
\label{compositing}
\hat{I_t} = \sum_{i=1}^{N}(c_{i}^{\prime}\alpha_{i}^{\prime} \prod_{j=i+1}^{N}(1-\alpha_{j}^{\prime})) .
\end{equation}
This rendering equation is completely differentiable, 
so our model can be trained from end-to-end.

\subsection{Loss Function}
Our overall loss combines an adaptive-bins loss to constrain the distribution of MPI according to each scene image and a synthesis loss to guide the network to synthesize images following the target views images.

\vspace{2mm}
\noindent\textbf{Adaptive-bins loss.} This loss term enforces that the distribution of MPI follows the ground truth value of the adaptive depth for each image:
\begin{equation}
\label{bin-loss}
L_{ada} = \sum_{x\in X} \min_{y \in p(b_i)}\lVert{x-y}\rVert^{2} + \sum_{y\in p(b_i)} \min_{x \in X}\lVert{x-y}\rVert^{2} \,,
\end{equation}
where $p(b_i)$ denotes the arranged depth of MPI and the set of all depth values in the ground truth image as $X$.

\vspace{2mm}
\noindent\textbf{Synthesis loss.} This loss aims at matching the synthesized target image with the ground truth by measuring the mean square error of RGB value and SSIM value~\cite{wang2004image}:
\begin{equation}
\label{syn-loss}
L_{syn} = \frac{1}{HW}\sum \left| \hat{I_t} - I_t \right| - SSIM(\hat{I_t}, I_t) \,,
\end{equation}
where $\hat{I_t}$ and $I_t$ are the synthesized novel image and ground truth image with the same size of $H \times W$.

The total loss is given by:
\begin{equation}
\label{total-loss}
L = \lambda_{ada}L_{ada} + L_{syn} \,,
\end{equation}
where $\lambda_{ada}$ is the parameter to balance the loss terms. 


\begin{table*}[!t]
  \caption{
  {\textbf{Overall comparison of SAMPLING with existing state-of-the-art approaches for novel view synthesis on the KITTI city dataset.} Note that $\uparrow$ denotes higher is better and $\downarrow$ means otherwise. The symbol $\dagger$ denotes the need for per-scene optimization and we use the average over all scenes as the final score. To ensure fairness, we follow the settings of MPI~\cite{tucker2020single} and MINE~\cite{li2021mine} and show the results with $N=64$.}}
  \centering
  \setlength{\tabcolsep}{6mm}{
    \begin{tabular}{c|c|ccc}
    \hline
    \textbf{Methods} & \textbf{Supervision} & \textbf{PSNR $\uparrow$} & \textbf{SSIM $\uparrow$} & \textbf{LPIPS $\downarrow$} \\
    \hline \hline
    NRW$\dagger$~\cite{martin2021nerf} & RGB + Point Clouds + Depth & 18.02 & 0.568 & 0.310  \\
    NPBG$\dagger$~\cite{aliev2020neural} & RGB + Point Clouds & 19.58 & 0.627 & 0.248  \\
    READ$\dagger$~\cite{li2022read} & RGB + Point Clouds & 23.48 & 0.781 & 0.132  \\
    Synsin~\cite{wiles2020synsin} & RGB + Point Clouds & 16.70 & 0.520 & -  \\
    PixelSynth~\cite{rockwell2021pixelsynth} & RGB + Point Clouds + Depth & 17.13 & 0.602 & -  \\
    3D-Photo~\cite{shih20203d} & RGB + Depth + Edges & 18.39 & 0.742 & 0.175  \\
    \hline
    LSI~\cite{tulsiani2018layer} & RGB & 16.52 & 0.572 & -  \\
    Deepview~\cite{flynn2019deepview} & RGB & 17.28 & 0.716 & 0.196  \\
    MPI~\cite{tucker2020single} & RGB & 19.54 & 0.733 & 0.158  \\
    MINE~\cite{li2021mine} & RGB & 21.65 & 0.818 & 0.117  \\
    \hline
    \textbf{SAMPLING (Ours)} & RGB & \textbf{23.67} & \textbf{0.883} & \textbf{0.101} \\
    \hline
    \end{tabular}%
    }
  \label{compare_SOTA}%
\end{table*}%

\section{Experiments}

%
We present quantitative and qualitative evaluations of our method on the KITTI~\cite{geiger2013vision} dataset and generalization performance on Tanks and Temples (T\&T)~\cite{knapitsch2017tanks}, compared with prior view synthesis methods. To assess the quality of the synthesized novel views, we mainly focus on the evaluation metrics of Peak Signal-to-Noise Ratio (PSNR), Structural Similarity Index Measure (SSIM)~\cite{wang2004image}, and Learned Perceptual Image Patch Similarity (LPIPS)~\cite{zhang2018unreasonable}. All metrics are computed over all pixels.




\begin{figure*}[htbp]
  \centering
  \includegraphics[width=15.0cm]{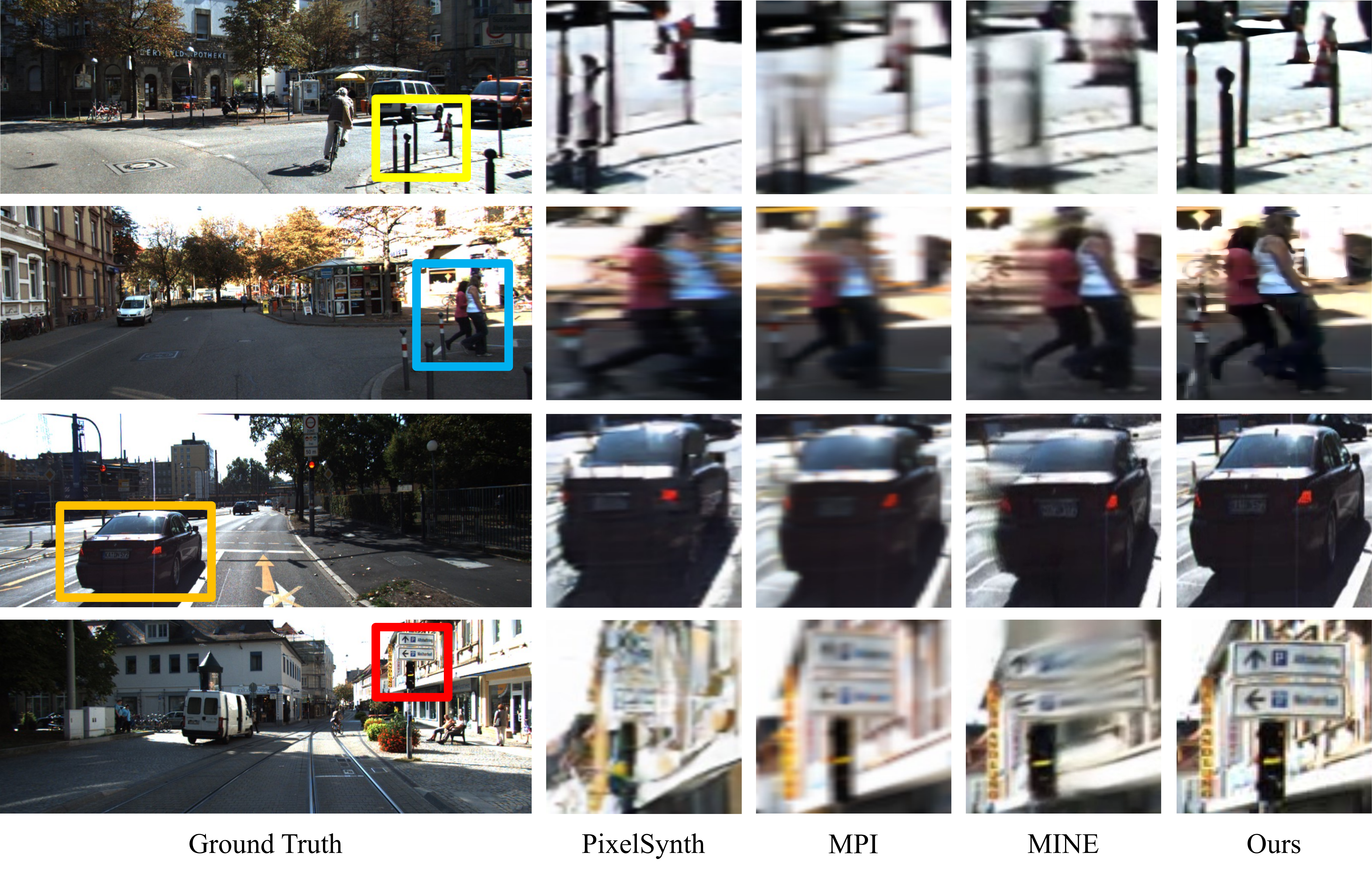}
  \caption{\textbf{Qualitative comparison of novel view synthesis on the KITTI dataset.} Visualization results show our method generates better details compared to other single-view NVS methods, including PixelSynth~\cite{rockwell2021pixelsynth}, MPI~\cite{tucker2020single}, and MINE~\cite{li2021mine}.}\label{fig:visual2}
  \vspace{-5pt}
\end{figure*}

%
%


\begin{table}[!t]
  \centering
  \caption{{\textbf{Generalization study on T\&T.} We evaluate the generalization of our method on the Tanks and Temples (T\&T) dataset that provides different scenes from KITTI.}}
  \setlength{\tabcolsep}{1.3mm}{
  \resizebox{0.46\textwidth}{!}{
    \begin{tabular}{cc|ccc}
    \hline
    \multirow{2}{*}{\textbf{Methods}} & \multirow{2}{*}{\textbf{Training Set}} & \multicolumn{3}{c}{T\&T} \\
    \cline{3-5}
    & & \textbf{PSNR} $\uparrow$ & \textbf{SSIM} $\uparrow$ & \textbf{LPIPS} $\downarrow$ \\
    \hline \hline
    NeRF~\cite{mildenhall2021nerf} & \multirow{5}{*}{T\&T} & 22.14 & 0.676 & -  \\
    NerfingMVS~\cite{wei2021nerfingmvs} &  & 19.31 & 0.464 & -  \\
    Monosdf~\cite{yu2022monosdf} &  & 21.48 & 0.689 & -  \\
    ResNeRF~\cite{xiao2022resnerf} &  & 23.39 & 0.795 & -  \\
    3D-Photo~\cite{shih20203d} &  & \textbf{23.63} & 0.848 & 0.136 \\
    \hline 
    MPI\*~\cite{tucker2020single} & \multirow{3}{*}{KITTI} & 18.62 & 0.614 & 0.260  \\
    MINE\*~\cite{li2021mine} & & 21.04 & 0.748 & 0.196  \\
    \textbf{Ours\*} & & 23.56 & \textbf{0.852} & \textbf{0.125} \\
    \hline
    \end{tabular}%
    }
    }
  \label{generalization-1}%
  \vspace{-10pt}
\end{table}%


\subsection{Evaluating Quality}
To demonstrate the efficacy of our method, we compare it to state-of-the-art methods for novel view synthesis. Following the settings of~\cite{tucker2020single, eigen2014depth}, we train our model on the city subset of the raw KITTI dataset, randomly taking either the left or the right image as the source (the other being the target) at each training step. 
Following~\cite{li2021mine, tucker2020single}, we evaluate the model on 4 test sequences of KITTI, cropping 5\% from all sides of all images.

We compare our method with state-of-the-art approaches for NVS using different types of 3D representations, including the traditional NeRF-based method~\cite{martin2021nerf}, Neural Point-based methods~\cite{aliev2020neural, li2022read}, generative model-based methods~\cite{wiles2020synsin, rockwell2021pixelsynth, shih20203d}, layer representation such as LDI-based method~\cite{tulsiani2018layer}, and MPI-based methods~\cite{flynn2019deepview, tucker2020single, li2021mine}. Note that the traditional NeRF-based and Neural Point-based methods require per-scene optimization and pre-processing for exploiting additional supervision.
Quantitative comparison results are presented in \tabref{compare_SOTA}. 



\begin{figure*}[htbp]
  \centering
  \includegraphics[width=16.5cm]{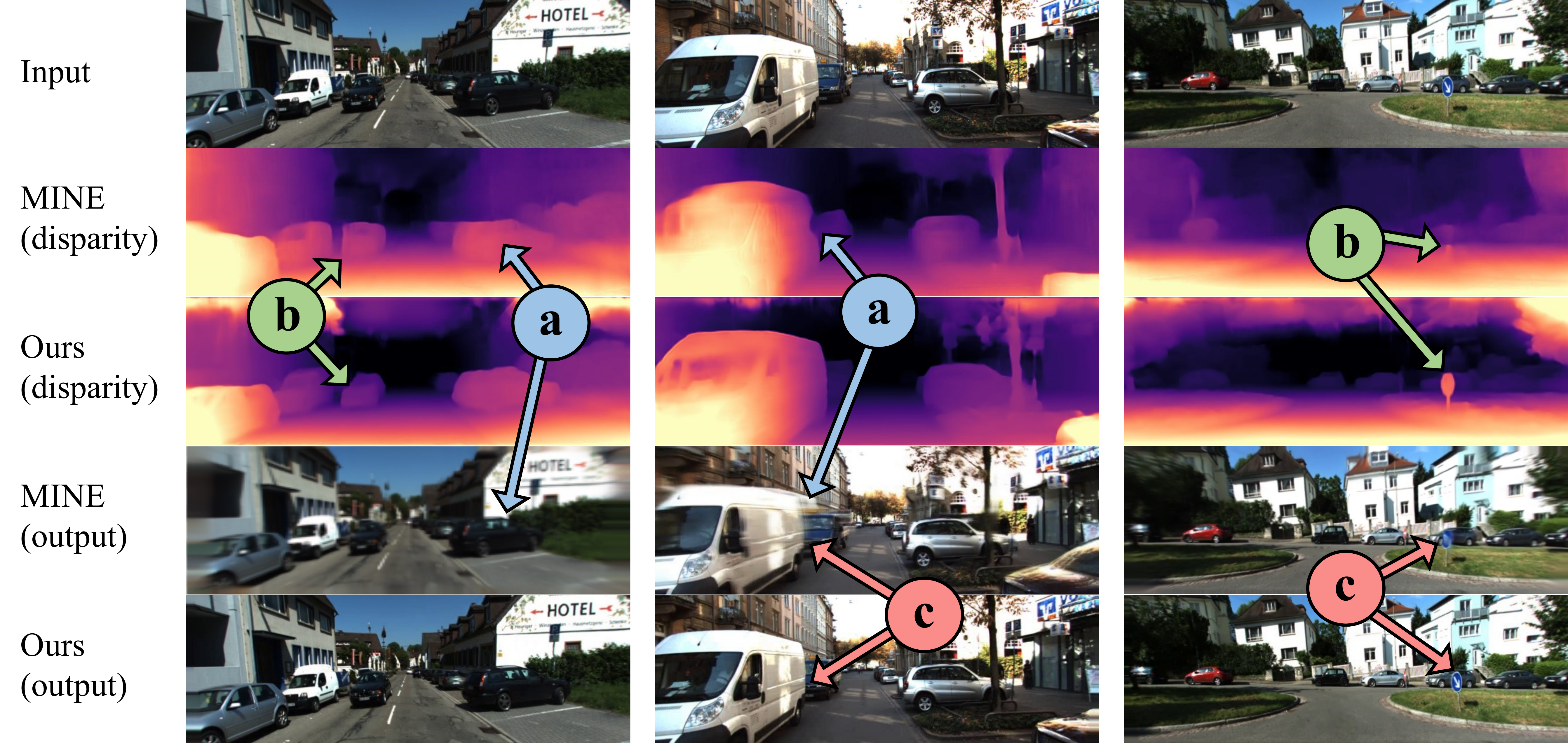}
  \caption{\textbf{Qualitative comparison of disparity map and novel view synthesis on the KITTI dataset.} (a) Disparity maps in~\cite{li2021mine} exhibit structural biases and missing objects, leading to unpleasant artifacts and distortions in the output. (b) The comparative disparity maps show that our method is capable of better recovering the spatial structure of complex scenes and intricate object boundaries. (c) Our method consistently delivers higher-quality and flawlessly disparity maps and outputs, even in challenging regions.}\label{fig:depth_comp3}
  \vspace{-5pt}
\end{figure*}

\vspace{2mm}
\noindent\textbf{Compared with NeRF-based methods.} We observe that our method outperforms NRW~\cite{martin2021nerf} by a large margin, although NRW introduces multiple supervision as well as paired poses to guarantee the training of the MLP. NPBG~\cite{aliev2020neural} and READ~\cite{li2022read} are two Neural Point-based methods, exploiting the extra point clouds for supervision and synthesizing large-scale driving scenes with neural rendering. Our method achieves competitive results across all three metrics and improves the SSIM to 0.883 compared with the state-of-the-art methods on KITTI. 

\vspace{2mm}
\noindent\textbf{Compared with generative models.} Based on generative models, SynSin~\cite{wiles2020synsin} and PixelSynth~\cite{rockwell2021pixelsynth} both utilize a high-resolution point cloud representation of learned features. 3D-Photo~\cite{shih20203d} presents a learning-based inpainting model combined with a Layered Depth Image, using depth and linked depth edges as additional supervision. Although these methods perform well in indoor scenes, they struggle with complex unbounded outdoor scenes due to the absence of strict geometric constraints and multi-scale features.

\vspace{2mm}
\noindent\textbf{Compared with layered representation methods.} Similar to MPI, LSI~\cite{tulsiani2018layer} applies a layer-structured 3D representation of a scene from a single input image. Compared with LSI, our method boosts the results by 7.15 on PSNR. DeepView~\cite{flynn2019deepview}, MPI~\cite{tucker2020single}, and MINE~\cite{li2021mine} are MPI-based or MPI-NeRF methods for novel view synthesis. Notably, we improve the performance of MPI for outdoor scenes on all metrics, compared with these existing methods. 

We also visually compare the view synthesis results in \figref{fig:visual2}. Our method produces more realistic images with high-quality details, more complete edge geometries, and fewer artifacts and distortions.
For small objects (\textit{e.g.}, pedestrians and traffic cones) and scene text (\textit{e.g.}, traffic signs), our method also shows favorable synthesis performance. The visualization confirms the effectiveness of our method in modeling the complex geometry and texture details of unbounded outdoor scenes.

We further show a qualitative comparison of disparity maps on the KITTI dataset in \figref{fig:depth_comp3}. Similar to~\cite{tucker2020single, li2021mine}, we use the models trained with KITTI to synthesize disparity maps from MPIs generated by our method and MINE~\cite{li2021mine}.
We can observe that MINE~\cite{li2021mine} displays missing and distorted areas in depth maps, leading to unpleasant visual artifacts. In contrast, our method excels in adaptively aligning the depth of various outdoor scenes, promoting the synthesis of more precise geometric shapes and well-aligned boundaries of visible objects. The proposed hierarchical refinement branch also serves as guidance for generating smooth and refined disparity maps, as well as synthesized outputs.
More results and videos are available in the supplementary material.

\begin{figure*}
  \centering
  \includegraphics[width=16.0cm]{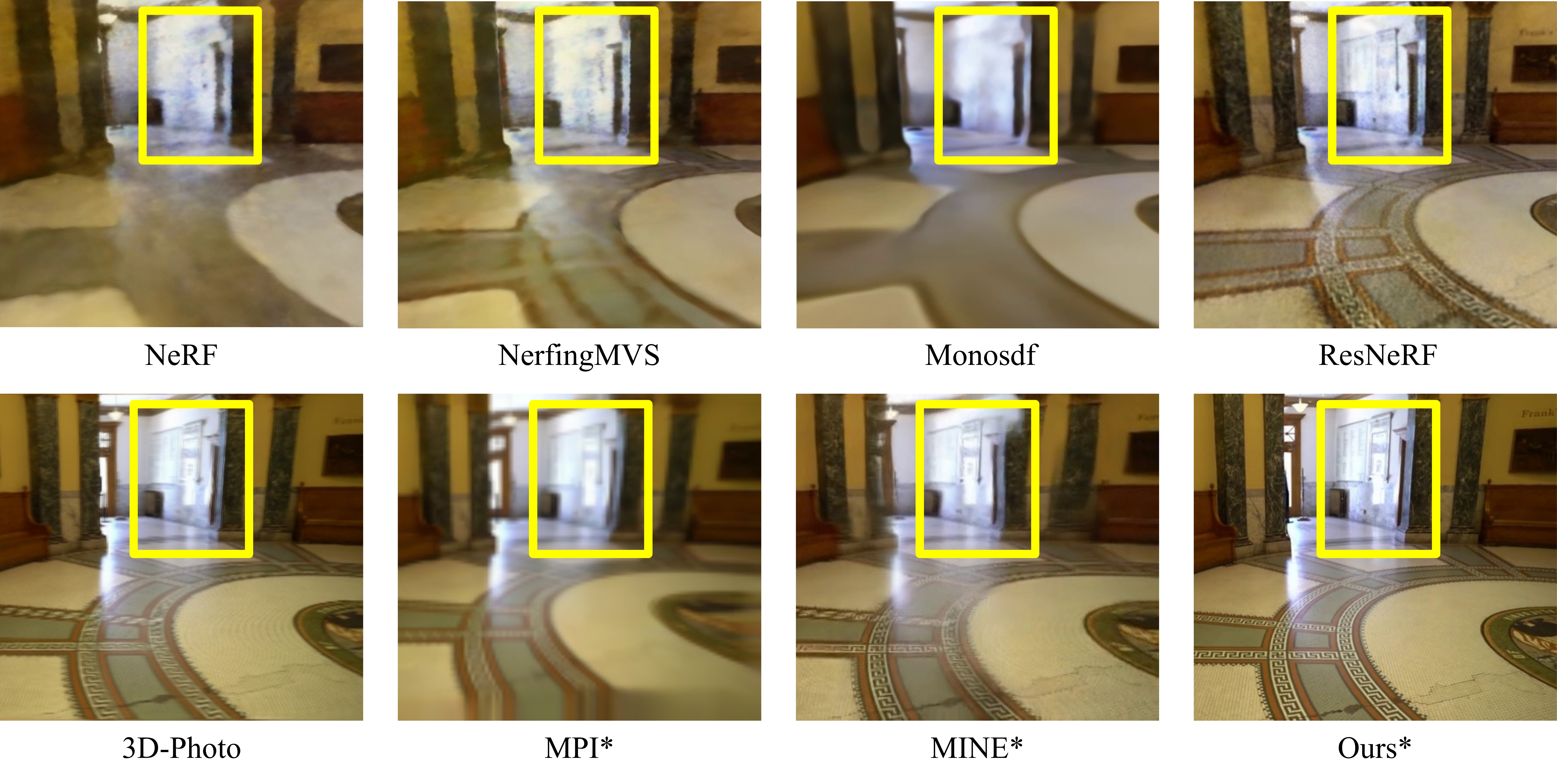}
  \caption{\textbf{The qualitative results of our method generalize to unseen dataset (T\&T).} The symbol ${\ast}$ denotes the model is trained on KITTI and evaluated on T\&T.}\label{fig:generalization}
  \vspace{-5pt}
\end{figure*}

\subsection{Generalization}
We further examine the generalization ability of our method using Tanks and Temples (T\&T) dataset. 
Specifically, we train our model on the KITTI dataset and evaluate it on the advanced sets of T\&T that contain indoor scenes.

%
We compare our model with existing methods for indoor scenes synthesis, such as NeRF~\cite{mildenhall2021nerf}, NerfingMVS~\cite{wei2021nerfingmvs}, Monosdf~\cite{yu2022monosdf}, and ResNeRF~\cite{xiao2022resnerf}. These methods employ explicit or implicit representation techniques to model a single scene with dense views as inputs. Note that these methods need to be trained separately for each scene, while our method can be trained in all scenes at once.
We also compare our method with 3D-Photo~\cite{shih20203d} as well as MPI-based methods, including MPI~\cite{tucker2020single} and MINE~\cite{li2021mine}. The quantitative results are presented in \tabref{generalization-1} and synthesis views of T\&T are shown in~\figref{fig:generalization}.

When evaluated on T\&T, our method still maintains a high level of performance. Quantitative results demonstrate that our method outperforms implicit representation approaches (\textit{e.g.}, Monosdf~\cite{yu2022monosdf} and ResNeRF~\cite{xiao2022resnerf}), despite their use of multiple dense views as input. 
3D-Photo~\cite{shih20203d} exploits a multi-layer representation for novel view synthesis and achieves state-of-the-art performance.  
Our method catches up with 3D-Photo~\cite{shih20203d} on PSNR and exceeds it in terms of SSIM and LPIPS.
Due to the relatively tight geometric constraints and different depth distributions of the interior, it can be observed that MPI-based methods~\cite{tucker2020single, li2021mine} show a certain degree of decline on T\&T, using the trained model on outdoor scenes (\textit{e.g.}, KITTI). Nevertheless, SAMPLING still exhibits good performance with minimal degradation. This can potentially be attributed to the employing of adaptive-bins MPI, leading to the image-level scene representation adaption. Additionally, our proposed hierarchical refinement branch aids in obtaining the multi-scale details of scenes, enhancing the generalization capability.


\begin{table}[!t]
  \centering
  \caption{{\textbf{Comparison of the proposed method with varying numbers of planes on the KITTI dataset.} $N$ denotes the preset number of planes for MPI.
  }}
  \setlength{\tabcolsep}{5.3mm}{
  \resizebox{0.46\textwidth}{!}{
    \begin{tabular}{c|ccc}
    \hline
    \bm{$N$} & \textbf{PSNR $\uparrow$} & \textbf{SSIM $\uparrow$} & \textbf{LPIPS $\downarrow$} \\
    \hline \hline
    8 &  20.02 & 0.812 & 0.139  \\
    16 & 22.18 & 0.856 & 0.114  \\
    32 & 23.53 & 0.879 & 0.105  \\
    64 & 23.67 & 0.883 & 0.101  \\
    \textbf{128} & \textbf{23.68} & \textbf{0.885} & \textbf{0.100} \\
    \hline
    \end{tabular}%
    }
    }
  \label{numebrs MPI}%
  \vspace{-8pt}
\end{table}%

\begin{table}[!t]
  \centering
  \caption{\textbf{Comparison of different strategies for MPI distribution.} Uniform-Fix and Log-Fix are two strategies for arranging MPI, both of which employ a static method for generating MPI and sampling.}
  \setlength{\tabcolsep}{1.7mm}{
  \resizebox{0.46\textwidth}{!}{
    \begin{tabular}{c|ccc}
    \hline
    \textbf{Variant} & \textbf{PSNR $\uparrow$} & \textbf{SSIM $\uparrow$} & \textbf{LPIPS $\downarrow$} \\
    \hline \hline
    Uniform-Fix MPI &  21.98 & 0.837 & 0.118  \\
    Log-Fix MPI & 22.53 & 0.862 & 0.112  \\
    \textbf{Adaptive-bins MPI (Ours)} & \textbf{23.67} & \textbf{0.883} & \textbf{0.101}  \\
    \hline
    \end{tabular}%
    }
    }
  \label{adaptive}%
  \vspace{-8pt}
\end{table}%

\subsection{Ablation Studies}
In this section, we conduct ablation experiments to analyze the effectiveness of each setting of our method, including the main components and hyperparameters.
We evaluate our method on KITTI dataset in the following experiments.

\vspace{2mm}
\noindent\textbf{Number of MPI planes.}
The performance of MPI representation is related to the number $N$ of planes. 
To study the influence of the number of MPI, we train our network with various values of $N$ and report results in \tabref{numebrs MPI}. 
We can see consistent improvements with increasing $N$ in our method. 
As the number of MPI increases, they can represent more complex scenes with a wider range of depth values. 
By contrast, sparse MPI (\textit{e.g.}, 8 planes) settings can lead to inadequate scene representation for large-scale outdoor scenes with a wide depth range. 
We use 64 planes of MPI in our experiments, which achieves good performance and computation cost trade-off.
%

\vspace{2mm}
\noindent\textbf{Type of MPI distribution.}
We examine the performance of three different strategies for MPI distribution and sampling, \textit{i.e.}, Uniform-Fix, Log-Fix, and proposed Adaptive-bins MPI. 
In our experiment, we replace the Adaptive-bins MPI module with the uniform-fix or log-fix strategy without changing the other modules.
Uniform-Fix MPI is a classical strategy employed by most MPI-based methods, such as MPI~\cite{tucker2020single} and MINE~\cite{li2021mine}. 
It divides the depth range at fixed intervals and randomly samples on MPI. 
Log-Fix MPI introduces a priori for the depth distribution and distributes the MPI according to the depth range in a log scale.
As shown in \tabref{adaptive}, compared with the two strategies, our Adaptive-bins MPI achieves optimal results by employing the adaptive bin distribution strategy per image, leading to a more efficient representation for unbounded outdoor scenes. 
%


\begin{table}[!t]
  \centering
  \caption{\textbf{Ablation study on network design.} Ada-bins stands for the Adaptive-bins MPI module. HRB is the abbreviation for Hierarchical Refinement Branch.}
  \setlength{\tabcolsep}{3.3mm}{
  \resizebox{0.46\textwidth}{!}{
    \begin{tabular}{c|ccc}
    \hline
    \textbf{Methods} & \textbf{PSNR $\uparrow$} & \textbf{SSIM $\uparrow$} & \textbf{LPIPS $\downarrow$} \\
    \hline \hline
    w/o Ada-bins & 21.98 & 0.837 & 0.118 \\
    w/o HRB & 22.87 & 0.869 & 0.109  \\
    w/o $L_{ada}$ & 22.25 & 0.858 & 0.113  \\
    \hline
    \textbf{Ours} & \textbf{23.67} & \textbf{0.883} & \textbf{0.101} \\
    \hline
    \end{tabular}%
    }
    }
  \label{ablation}%
  \vspace{-5pt}
\end{table}%

\vspace{2mm}
\noindent\textbf{Effectiveness of Each Module.}
We further investigate how each proposed module contributes to the final performance. We first verify the effectiveness of Adaptive-bins MPI by removing this module and exploiting uniform-fixed MPI followed by random sampling. We report the results in \tabref{ablation}. The results show that the Adaptive-bins MPI module plays a key role in modeling the entire outdoor scenes and generating more efficient representations. Then, we remove the hierarchical refinement branch, proving its usefulness in improving image quality and capturing texture details.
Moreover, experimental results indicate that introducing the adaptive-bins loss function helps with better distributing planes according to each image. Qualitatively, our method achieves favorable results with the overall combination of each module.

\begin{figure}[h]
  \centering
  \includegraphics[width=8.2cm]{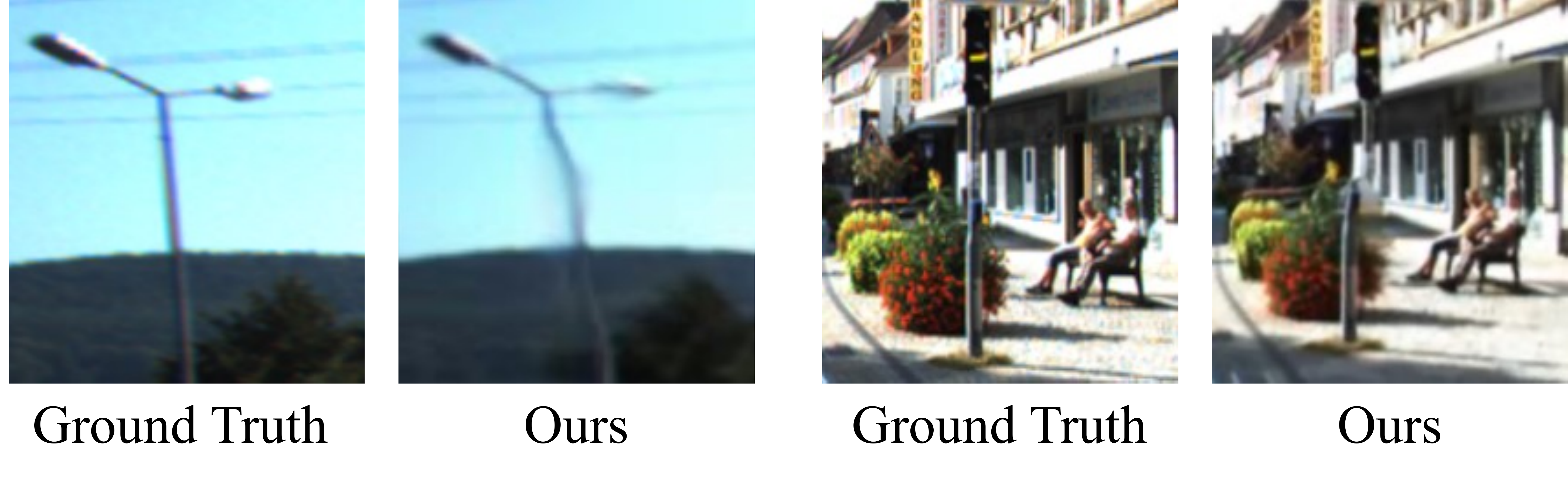}
  \caption{\textbf{Failure cases.} Due to the extremely narrow geometries (\eg, street light pole) and inhomogeneous diffuse reflections, our method fails in modeling these areas and generates images distorted and misaligned.}\label{fig:failure case}
  \vspace{-5pt}
\end{figure}

\subsection{Limitations \& Failure Cases}
Our method is based on MPI representation and, as a result, inherits certain limitations. When the synthesis view is significantly distant from the observation view, the generated images have relatively obvious visual distortions and artifacts. As with other MPI-based methods, the areas with strong diffuse light and slender geometric shapes may lead to distorted representation in planes as well as rendered output, as shown in~\figref{fig:failure case}. Learning how to synthesize images in these hard cases could be a promising research topic.

\section{Conclusion}
In this paper, we present SAMPLING, an improved MPI-based novel view synthesis method from a single-view image for outdoor scenes.
%
To address the difficulty of representing intricate geometries in unbounded outdoor scenes, we introduce Adaptive-bins MPI, which can adaptively distribute the planes of MPI in different depths for each scene image. 
Besides, we propose a Hierarchical Refinement Branch to fuse multi-scale information for better image detail generation. 
%
%
%
%
%
Experiment results show that our method enhances the efficiency and quality of MPI representation, especially in modeling complex geometries and high-frequency details.
Our method achieves new state-of-the-art view synthesis results on the large-scale outdoor dataset.
Furthermore, experimental results show that our method has a strong generalization ability on unseen scenes.

\section*{Acknowledgements}
This work was supported by National Key R\&D Program of China (Grant No. 2022ZD0160305). This work was also a research outcome of Key Laboratory of Science, Technology and Standard in Press Industry (Key Laboratory of Intelligent Press Media Technology).
\vspace{3mm}

 {\small
 \bibliographystyle{unsrt}
 \bibliography{egbib}
 }

\end{document}